\documentclass[10pt,letterpaper]{article}
\usepackage[a4paper,
            bindingoffset=0.2in,
            left=1in,
            right=1in,
            top=1in,
            bottom=1in,
            footskip=.25in]{geometry}
\usepackage{tabularx,booktabs}
\usepackage{url}
\usepackage{fullpage}
\usepackage{graphicx}
\usepackage{graphics}
\usepackage{lipsum}
\usepackage{amsmath}
\usepackage{sidecap}
\usepackage{caption}
\usepackage{authblk}
\usepackage{times}
\usepackage{float}
\usepackage{setspace}
\usepackage[dvipsnames,svgnames,x11names]{xcolor}
\usepackage{cite}
\usepackage[ruled,linesnumbered]{algorithm2e}

\DeclareCaptionLabelFormat{cont}{#1~#2\alph{ContinuedFloat}}
\usepackage{subfiles}
\newcommand{\junk}[1]{}
\usepackage{caption}
\usepackage[colorlinks,linkcolor=blue]{hyperref}

\newcounter{extendedfigure}

\DeclareCaptionType{extendedfigure}[Supplementary Figure][Extended Data Figures]

\begin{document}
\thispagestyle{empty}
\setlength{\baselineskip}{12pt}

\title{MedKGent: A Large Language Model Agent Framework for Constructing Temporally Evolving Medical Knowledge Graph}
\newcommand{\subheading}[1]{\vspace{0.5em}\noindent\textbf{#1.}}

\author[1]{Duzhen Zhang$^*$}
\author[1]{Zixiao Wang$^*$}
\author[2]{Zhong-Zhi Li$^*$}

\author[3]{Yahan Yu}
\author[2]{Shuncheng Jia}
\author[1]{Jiahua Dong}
\author[4]{Haotian Xu}
\author[2]{Xing Wu}
\author[5]{Yingying Zhang}
\author[6]{Tielin Zhang}
\author[7]{Jie Yang}

\author[1]{Xiuying Chen$^\dagger$} 
\author[1,8]{Le Song$^\dagger$}

\affil[1]{\small Mohamed bin Zayed University of Artificial Intelligence, Abu Dhabi, UAE}
\affil[2]{\small University of Chinese Academy of Sciences, Beijing, China}
\affil[3]{\small Kyoto University, Kyoto, Japan}
\affil[4]{\small Tsinghua University, Beijing, China}
\affil[5]{\small East China Normal University, Shanghai, China}
\affil[6]{\small Center for Excellence in Brain Science and Intelligence Technology, Chinese Academy of Sciences, Shanghai, China}
\affil[7]{\small Brigham and Women’s Hospital, Harvard Medical School, Boston, MA, USA}
\affil[8]{\small GenBio AI, San Francisco, USA}

\affil[$*$]{Contribute equally.}
\affil[$\dagger$]{Corresponding author. Email: xiuying.chen@mbzuai.ac.ae; le.song@mbzuai.ac.ae.}

\date{}
\maketitle
\vspace{-0.5em} 
\begin{abstract}

The rapid expansion of medical literature challenges the scalable structuring of domain knowledge. 
Knowledge Graphs (KGs) offer a solution, yet current construction methods lack generalizability and ignore the temporal dynamics of evolving knowledge. 
To address this, we introduce MedKGent, a Large Language Model (LLM) agent framework for building temporally evolving medical KGs.
Using over 10 million PubMed abstracts from 1975 to 2023, MedKGent incrementally constructs a KG daily via two specialized agents. 
The Extractor Agent identifies knowledge triples and assigns confidence scores, while the Constructor Agent integrates these triples into a temporal graph, reinforcing recurring knowledge and resolving conflicts. 
The resulting KG contains 156,275 entities and 2,971,384 triples, making it, to our knowledge, the largest LLM-derived medical KG to date. 
Automated and expert assessments showed triple-validity rates approaching 90\%. 
In downstream evaluations, MedKGent-KG significantly improved retrieval-augmented generation for five LLMs across seven medical question-answering benchmarks. 
Together, these results position MedKGent as a scalable and temporally aware infrastructure for medical knowledge representation and literature-grounded AI research.

\end{abstract}

\newpage
\section{Introduction}

The rapid expansion of medical literature presents mounting challenges for organizing, synthesizing, and accessing structured domain knowledge \cite{eberendu2016unstructured}.
With millions of publications spanning decades and disciplines, clinicians and researchers increasingly struggle to keep pace with new findings, reconcile conflicting evidence, and extract actionable insights from an ever-growing body of unstructured text \cite{jin2024pubmed}. 
These challenges underscore the need for scalable approaches that can transform unstructured medical literature into structured representations to support data-driven discovery and literature-grounded AI reasoning \cite{wang2025biomedical,aftab2024optimizing}.

Knowledge Graphs (KGs) offer a powerful paradigm for organizing biomedical knowledge by representing entities and their relations as relational triples \cite{li2020real,yu2022bios}.
By encoding semantic content and contextual connections in a graph-based form, KGs enable machine reasoning, large-scale knowledge integration, and interpretable downstream analysis \cite{chen2020review,hogan2021knowledge}. 
They have been widely applied to a range of biomedical research and informatics tasks, including drug repurposing studies, disease-gene association analysis, and knowledge-augmented biomedical question answering \cite{nicholson2020constructing,gao2022kg,zheng2021pharmkg}. 
More recently, KGs have been increasingly integrated into Large Language Model (LLM) workflows, most notably via Retrieval-Augmented Generation (RAG), to provide structured external knowledge that improves factual accuracy, interpretability, and reasoning in knowledge-intensive medical applications such as Question Answering (QA) \cite{wu2024guiding}. 
Together, these developments position KGs as foundational infrastructure for structuring and leveraging biomedical knowledge at scale \cite{li2020kghc,ernst2015knowlife}.

Despite their promise, the automatic construction of medical KGs from unstructured literature remains challenging due to the scale, heterogeneity, and linguistic complexity of biomedical texts.
Early efforts largely relied on rule-based systems and feature-engineered machine learning pipelines for named entity recognition and relation extraction \cite{wu2023medical,cui2023review,li2009two,hong2020novel}.
Although these approaches established important foundations, they require substantial domain expertise, depend on fixed schemas, and are difficult to adapt to emerging relation types or rapidly evolving biomedical concepts \cite{abu2023healthcare}.
Subsequent advances in deep learning improved extraction performance by modeling richer syntactic and semantic patterns, but typically rely on large annotated datasets and task-specific supervision, limiting scalability and flexibility across domains \cite{lu2025biomedical,li2017neural,lee2020biobert,zhang2023constructing,zhang2025comprehensive}.

LLMs, exemplified by GPT-4 \cite{openai2023gpt4}, have recently emerged as a compelling alternative for KG construction \cite{ye2024construction,bai2025construction,yang2025graphusion,gilbert2024augmented,zhu2024llms,lairgi2024itext2kg,zhang2024extract,ningurbankgent,han2024pive,chen2024sac}. 
Pretrained on large-scale, heterogeneous corpora, LLMs support flexible, zero-shot information extraction with minimal supervision and can accommodate emerging relation types through prompt engineering without fixed schemas or retraining \cite{ding2024automated,dong2024survey}. 
In the biomedical and health domains, recent studies have explored LLM-based or LLM-assisted KG construction for literature-grounded disease analysis and downstream applications \cite{ying2024cortex}, including dynamic Alzheimer's disease graphs for QA \cite{li2024dalk}, ontology-assisted rare disease KGs \cite{cao2024automatic}, and ADHD-focused graph modeling \cite{otal2024new}. 
These studies reflect a rapidly developing field in which LLMs are increasingly used to reduce manual schema engineering and support flexible biomedical relation extraction. 
However, existing systems are often designed for specific diseases or task settings, and therefore do not fully address the need for a comprehensive, continuously updatable medical KG that captures how evidence accumulates, evolves, and conflicts over time.

Moreover, most existing LLM-based KG construction methods treat medical literature as a static corpus, aggregating extracted relations without explicit consideration of temporal provenance.
Such approaches obscure the inherently dynamic nature of medical knowledge, in which new evidence may refine, contradict, or reinforce prior findings.
In addition, confidence estimation is rarely incorporated \cite{gu2024probabilistic}, limiting the ability to resolve inconsistencies or prioritize robust, repeatedly supported knowledge.
Without modeling temporal progression or epistemic uncertainty, current methods fall short of producing coherent, trustworthy, and dynamically evolving representations of biomedical knowledge.

To address these limitations, we introduce MedKGent, an LLM agent framework for constructing a temporally evolving medical KG. 
We curated 10,014,314 PubMed abstracts published between January 1, 1975, and December 31, 2023, and organized them into a fine-grained daily time series to preserve the chronological trajectory of biomedical knowledge emergence. 
In contrast to static approaches, MedKGent processes abstracts incrementally on a day-by-day basis, enabling the KG to evolve dynamically while remaining extensible to future updates. 
The framework consists of two coordinated agents, both built on the open-source Qwen2.5-32B-Instruct model \cite{qwen2.5}. 
The Extractor Agent identifies relational triples from each abstract and assigns initial confidence scores via sampling-based confidence estimation \cite{wangself,taubenfeld2025confidence,chen2024universal}, which are used to filter low-confidence extractions and guide downstream processing. 
It also enriches extracted entities and relations with detailed attribute information to support downstream applications such as entity disambiguation and biomedical information retrieval. 
The Constructor Agent integrates the curated triples into a temporal graph through continual interaction with a graph database. 
Guided by confidence scores and timestamps, it incrementally refines the KG. 
Recurrent evidence strengthens the reliability of existing facts, while conflicting relations are resolved through accumulated evidence, thereby preserving coherence as the body of literature evolves.
The resulting KG contains 156,275 entities and 2,971,384 relational triples---representing, to our knowledge, the largest LLM-derived medical KG constructed to date.

To evaluate the quality of the resulting KG, we conducted both automated and expert assessments. 
Two State-Of-The-Art (SOTA) LLMs—GPT-4.1 \cite{openai_gpt4.1} and DeepSeek-v3 \cite{liu2024deepseek}, alongside two PhD-level domain experts independently evaluated the extracted triples. 
All assessments reported high accuracy, approaching 90\%, with strong agreement across evaluators, supporting the overall reliability of the KG.
To assess downstream utility, we evaluated knowledge-augmented question answering on seven medical QA datasets, including four widely used benchmarks, MMLU-Med, MedQA-US, PubMedQA*, and BioASQ-Y/N \cite{xiong2024benchmarking}, and three recently released differential diagnosis datasets: MedDDx-Basic, MedDDx-Intermediate, and MedDDx-Expert \cite{su2025kgarevion}. 
We compare three settings across five leading LLMs (GPT-4-turbo \cite{openai_gpt_turbo}, GPT-3.5-turbo \cite{openai_gpt_turbo}, DeepSeek-v3 \cite{liu2024deepseek}, Qwen-Max \cite{yang2025qwen3}, and Qwen-Plus \cite{yang2025qwen3}): direct answering without retrieval, RAG using an established biomedical KG (SemMedDB \cite{kilicoglu2012semmeddb}), and RAG using our KG constructed by MedKGent. 
Across all tasks and models, both knowledge-augmented settings improve over direct answering, with OurKG-based RAG consistently achieving the strongest performance, supporting its utility as a literature-grounded knowledge source for biomedical QA. Collectively, these findings establish MedKGent as a temporally informed and scalable infrastructure for biomedical knowledge representation, medical research, and AI-enabled discovery.

\section{Results}
  \subheading{Method overview}

\begin{figure}[h]
    \centering
    \includegraphics[width=1.0\textwidth]{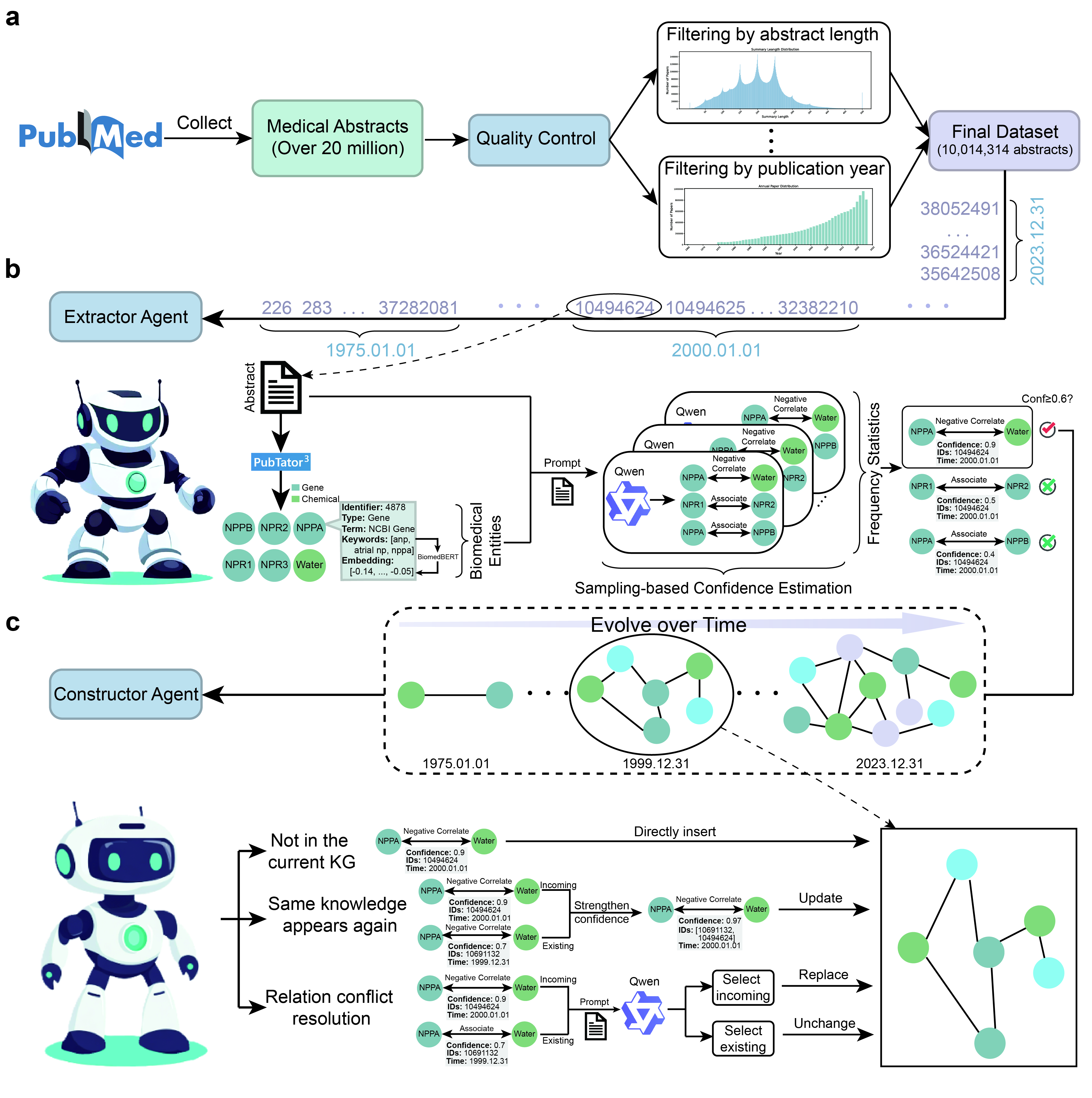}  
   \caption{Overview of the MedKGent framework for constructing a temporally evolving medical KG. \textbf{a}. Data collection and preprocessing pipeline. Over 20 million PubMed abstracts were retrieved and subjected to quality control, including filtering by abstract length and publication year, yielding 10,014,314 abstracts organized as a daily time series from 1975–2023 to enable fine-grained temporal analysis. 
    \textbf{b}. Entity and relation extraction by the Extractor Agent. Each biomedical abstract is processed using PubTator3 to detect and normalize entities across six biomedical categories. Entities are enriched with exact keywords and semantic embeddings, followed by LLM-based inference of biomedical relations with sampling-based confidence estimation; low-confidence triples are discarded. 
    \textbf{c}. Incremental graph construction by the Constructor Agent. High-confidence triples are integrated into the evolving KG, with confidence scores strengthened when re-encountered, PubMed IDs appended for provenance, and timestamps refreshed. Conflicting relations are resolved through LLM-based reasoning, ensuring consistent and temporally aware knowledge integration.}
    \label{fig:method_overview}
\end{figure}

As illustrated in Figure \ref{fig:method_overview} \textbf{a}, we collected over 20 million abstracts from \href{https://pubmed.ncbi.nlm.nih.gov/}{PubMed} as the data source for constructing the medical KG, given their concise and information-dense summaries of research findings. 
These abstracts underwent a series of quality control procedures, including filtering by abstract length (Supplementary Figure 5 \textbf{a}) and publication year (Supplementary Figure 5 \textbf{b}). 
The final dataset, comprising 10,014,314 abstracts, was organized into a daily time series from January 1, 1975, to December 31, 2023 (Supplementary Figure 5 \textbf{c}), enabling high-resolution temporal analysis.

We introduce MedKGent, an LLM-based agent framework designed to construct a temporally evolving medical KG. 
The framework processes biomedical abstracts sequentially from January 1, 1975, to December 31, 2023, enabling incremental KG growth while ensuring adaptability for future updates. 
MedKGent comprises two coordinated agents---the Extractor Agent and the Constructor Agent---deployed via a self-hosted API using the open-source Qwen2.5-32B-Instruct model \cite{qwen2.5}. 
As shown in Figure \ref{fig:method_overview} \textbf{b}, for a given biomedical abstract (\emph{e.g.}, PubMed ID 10494624), the Extractor Agent first utilizes the PubTator3 tool \cite{wei2024pubtator} to identify entities across six categories: Gene, Disease, Chemical, Variant, Species, and CellLine. 
This tool extracts entities, assigns types, and normalizes synonymous mentions with unique identifiers, facilitating entity disambiguation and retrieval---key steps for accurate KG construction. 
For example, ``NPPA'' is classified as a Gene with identifier ``4878'' and terminology ``NCBI Gene''. 
Moreover, the Extractor Agent enriches each entity with Exact Keywords (standardized to lowercase and mapped to a single identifier, \emph{e.g.}, aliases: ``anp'', ``atrial np'', and ``nppa'' for ``NPPA'') and Semantic Embedding attributes---a vector generated by BiomedBERT \cite{pubmedbert} to capture semantic context. 
These enriched representations enhance retrieval precision and efficiency, particularly when explicit identifiers are unavailable.

Next, the Extractor Agent uses the abstract and extracted entities to prompt the LLM, inferring semantic relationships between entity pairs (prompt shown in Supplementary Figure 6 \textbf{a}). 
We define 12 core biomedical relation types: seven bidirectional (Associate, Negative Correlate, Positive Correlate, Compare, Cotreat, Interact, Drug Interact) and five unidirectional (Cause, Inhibit, Treat, Stimulate, Prevent), as detailed in Supplementary Figure 6 \textbf{d}. These relations can be flexibly extended by prompt design, enabling MedKGent to incorporate emerging medical relations with minimal manual intervention, eliminating the need for rigid schemas or retraining typically required in supervised pipelines. 
Inspired by the self-consistency principle \cite{wangself, taubenfeld2025confidence, chen2024universal}, the Extractor Agent uses sampling-based confidence estimation to assign an initial confidence score to each triple. 
For a given extraction prompt, the Extractor Agent performs multiple parallel LLM inferences, calculating the frequency of each triple across outputs as its confidence score. 
For instance, if the triple (NPPA, Negative Correlate, Water) appears in 90\% of the outputs, its confidence score is 0.9. 
Low-confidence triples (score$<$0.6) are filtered out, and only high-confidence triples are retained for downstream graph construction. 
Each triple is also annotated with the PubMed ID of the source abstract and a timestamp, ensuring traceability and source attribution. 
For example, (NPPA, Negative Correlate, Water) would have a PubMed ID of 10494624 and a timestamp of 2000-01-01.

As shown in Figure \ref{fig:method_overview} \textbf{c}, for each retained triple, such as (NPPA, Negative Correlate, Water), the Constructor Agent checks its presence in the current KG. 
If absent (\emph{i.e.}, either the head or tail entities are missing), it is inserted; if present, its confidence score is strengthened according to Equation (\ref{eq:update_conf}).
The associated PubMed ID is appended, and the timestamp is updated to reflect the latest publication. 
For example, if an existing triple (NPPA, Negative Correlate, Water) has a confidence score of 0.7, PubMed ID 10691132, and timestamp 1999-12-31, and a new occurrence with a confidence score of 0.9, PubMed ID 10494624, and timestamp 2000-01-01 is encountered, the updated triple will have a confidence score of 0.97, PubMed IDs [10691132, 10494624], and a timestamp of 2000-01-01. 
If the head and tail entities are present but the relation differs, such as existing (NPPA, Associate, Water) vs. incoming (NPPA, Negative Correlate, Water), only the most appropriate relation is maintained. 
The Constructor Agent invokes the LLM to resolve the conflict by selecting the more suitable relation, considering both the existing and incoming triple's confidence scores and timestamps. 
If the LLM selects the new triple, the existing one is replaced; otherwise, no changes are made. 
The prompt design for relation conflict resolution is shown in Supplementary Figure 6 \textbf{c}. 
Together, the two agents extract structured medical facts and integrate them into a dynamic, time-aware KG. See more details in the Section \ref{method}.

\subheading{Structural Characterization of the Knowledge Graph}

In this section, we detail the structural characteristics of the medical KG we constructed, with an emphasis on the distribution of node types, relationship types, and the confidence scores of relationship triples. 
We also present a visualization of a subgraph centered on COVID-19 to illustrate the graph's structure.

Using the MedKGent framework, we extracted knowledge triples from the abstracts of 10,014,314 medical papers, with 3,472,524 abstracts (34.68\%) yielding extractable triples. 
The relatively low extraction rate can be attributed to several factors: first, some abstracts lacked sufficient structured information for triple extraction; second, only triples with a confidence score exceeding 0.6 were retained, excluding those with lower confidence; and third, some triples extracted by LLMs contained formatting issues, such as extraneous or irrelevant characters, which were discarded. 
In total, our Extractor Agent identified 8,922,152 valid triples from the abstracts. 
However, the extracted triples contained a significant number of duplicates and conflicts. 
To resolve this, our Constructor Agent integrates the triples in chronological order. 
During this process, duplicates are merged, with the confidence score for each triple increasing in proportion to its frequency, reflecting greater certainty. For conflicting triples, where the same entity pair is associated with multiple relations, the Constructor Agent retains the most appropriate relationship. Following this consolidation, the final KG comprises 2,971,384 distinct triples.

\begin{figure}[h]
    \centering
    \includegraphics[width=1.0\textwidth]{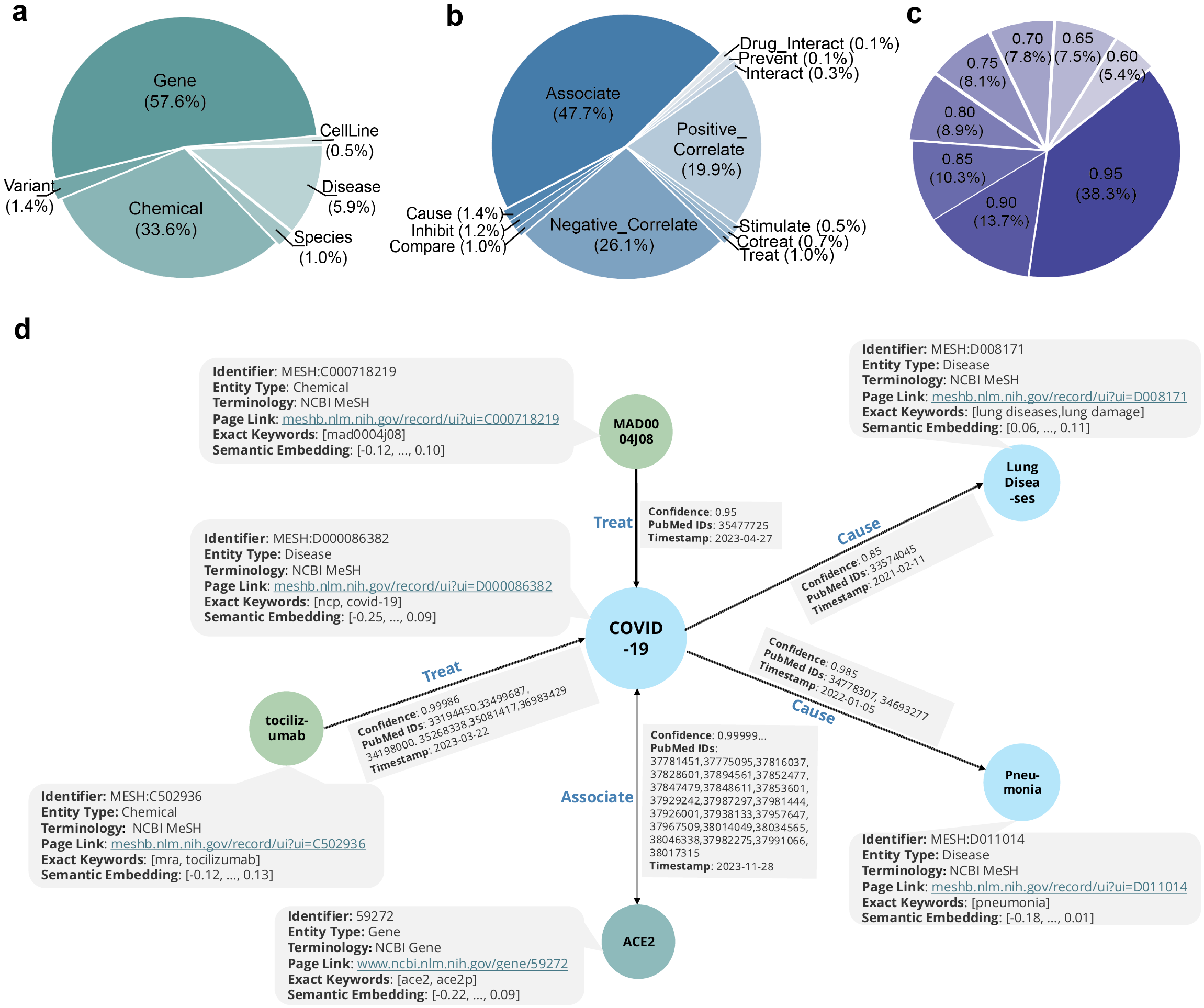}  
    \caption{A comprehensive statistical analysis and visualization of the constructed KG, consisting of 156,275 nodes and 2,971,384 relationship edges. \textbf{a}. Node distribution within the KG, with Gene and Chemical nodes predominating, and smaller proportions of Disease, Variant, Species, and CellLine. 
    \textbf{b}. Relationship type distribution within the KG, highlighting the prevalence of ``Associate'' relationships, followed by ``Negative Correlate'' and ``Positive Correlate'', with less common interactions such as ``Interact'', ``Prevent'', and ``Drug\_Interact''. 
    \textbf{c}. The distribution of confidence scores for relationship triples, discretized to the nearest smaller 0.05 increment, ensures values are rounded down to the closest multiple of 0.05. 
    This distribution reveals a clear dominance of high-confidence triples, particularly those with scores of 0.95, underscoring the robustness of the KG. \textbf{d}. Local subgraph visualization centered on COVID-19, displaying five surrounding relationship triples. 
    Each node is characterized by key attributes, including Identifier, Entity Type, Terminology, Page Link, Exact Keywords, and Semantic Embedding, facilitating efficient entity linking through exact or similarity matching. The relationships in the KG are further enriched by attributes such as Confidence, PubMed IDs, and Timestamp, enhancing traceability, accuracy, and temporal relevance.}
    \label{fig:kg_statis}
\end{figure}

We conducted a comprehensive statistical analysis of the final constructed KG, which comprises 156,275 nodes. 
As shown in Figure \ref{fig:kg_statis} \textbf{a}, the node distribution is predominantly dominated by Gene and Chemical nodes, with smaller proportions of other entities such as Disease, Variant, Species, and CellLine. 
The KG includes 2,971,384 relationship triples (edges), representing a range of interactions between entities, as illustrated in Figure \ref{fig:kg_statis} \textbf{b}. 
The most common relationship type is ``Associate'', followed by ``Negative Correlate'' and ``Positive Correlate'', indicating strong associations between medical entities. 
Less frequent relationships, such as ``Interact'', ``Prevent'', and ``Drug\_Interact'', provide additional insights into the complexities of medical interactions. 
The distribution of confidence scores for these relationship triples, shown in Figure \ref{fig:kg_statis} \textbf{c}, with confidence values discretized to the nearest smaller 0.05 increment (rounding down to the closest multiple of 0.05), reveals a clear dominance of high-confidence triples. 
A significant proportion of triples exhibit confidence scores of 0.95, reflecting the cumulative increase in confidence resulting from the repetition of triples during the graph construction process. 
This high-confidence distribution reinforces the reliability and robustness of the KG.

We visualized a local subgraph of the constructed KG with COVID-19 as the central node, highlighting five surrounding relationship triples, as shown in Figure \ref{fig:kg_statis} \textbf{d}. 
Each node is characterized by six key attributes: 
the Identifier, which uniquely references the node and normalizes multiple synonymous mentions to a standardized terminology entry; 
the Entity Type, which classifies the entity; 
the Terminology, which maps the entity type to its corresponding standard terminology; 
the Page Link, providing a reference to the entity in the Terminology; 
the Exact Keywords, which lists common names and aliases of the entity in lowercase; 
and the Semantic Embedding, a vector representation of the entity. 
In practice, these attributes facilitate entity linking within a query by matching entities to their corresponding nodes in the KG. 
When the Identifier of an entity in the query is available, entity linking can be efficiently performed using this unique reference. 
In the absence of an Identifier, precise matching is achieved by checking whether the entity appears in the Exact Keywords list of a specific node. 
Alternatively, semantic vectors of the query entities can be compared with those in the KG to identify the most similar entities, enabling semantic similarity matching. 
This approach is particularly beneficial for entities with multiple names, ensuring accurate linking even when not all aliases are captured in the Exact Keywords list.

The relationships between entities are characterized by three key attributes. Confidence reflects the reliability of the relationship, with higher values indicating greater certainty based on its frequency across multiple sources. 
The PubMed IDs attribute lists the PubMed identifiers of the papers from which the relationship is derived, enabling easy access to the original publications via the \href{https://pubmed.ncbi.nlm.nih.gov/}{PubMed} website. 
If the relationship appears in multiple papers, all relevant PubMed IDs are included, further increasing the confidence score. 
Finally, Timestamp denotes the most recent occurrence of the relationship, specifically the publication date of the latest paper. 
Notably, while Timestamp captures only the latest appearance, the full temporal span of the relationship---including its earliest mention---can be readily retrieved through the associated PubMed IDs via the PubMed website. 
These attributes collectively enhance the traceability, accuracy, and temporal relevance of the relationships within the KG. 
Additional local subgraph visualizations are presented in Supplementary Figures 1-4.

\subheading{Quality Assessment of Extracted Relationship Triples}

To systematically evaluate the reliability of the Extractor Agent, which generated relationship triples from 3,472,524 abstracts, we conducted both automated and manual assessments to characterize extraction accuracy.

\begin{figure}[h]
    \centering
    \includegraphics[width=1.0\textwidth]{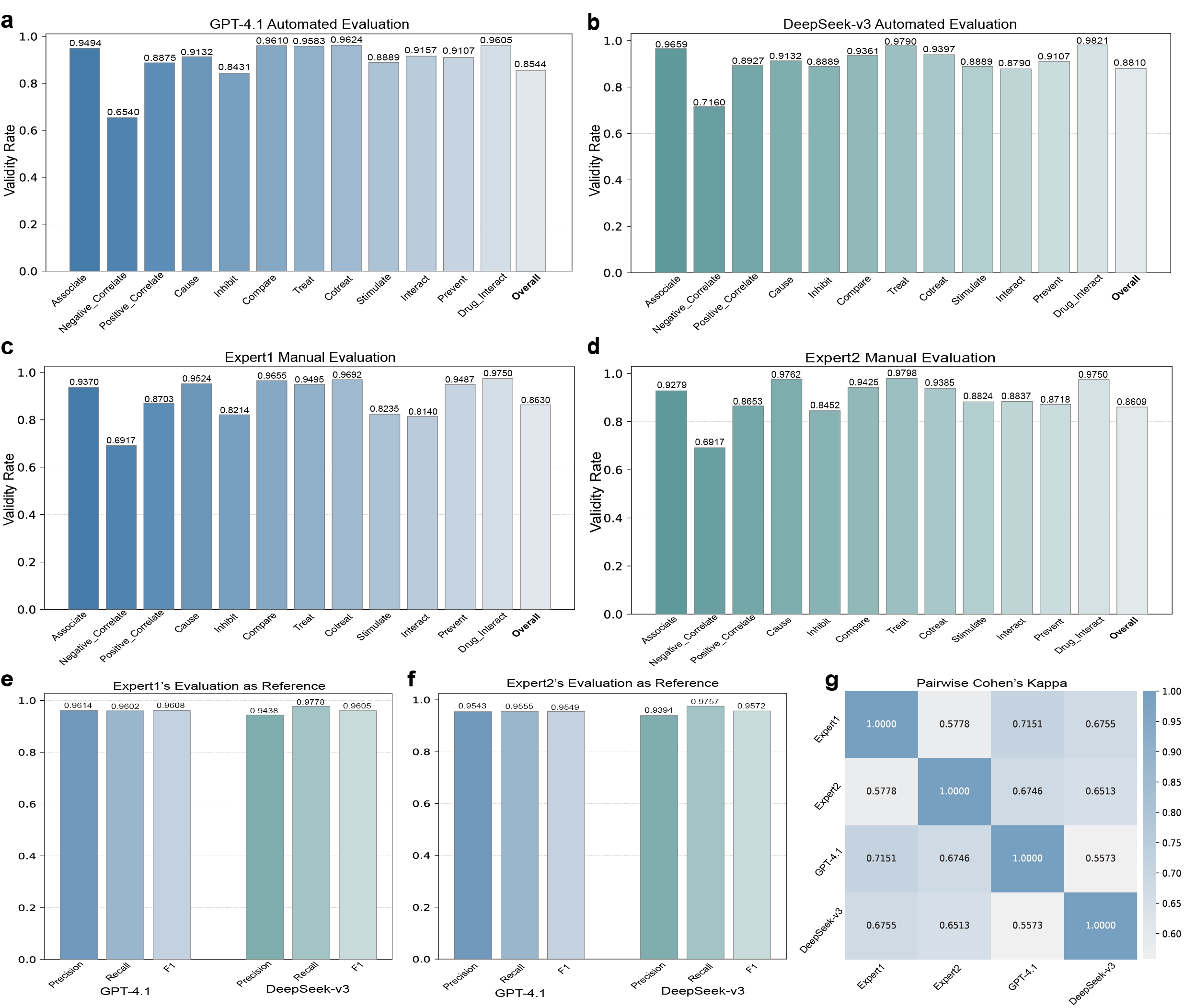}  
        \caption{Comprehensive evaluation of extraction quality for relationship triples generated by the Extractor Agent. 
    Systematic assessment of extraction accuracy using both automated evaluations by LLMs and independent manual expert review. 
    \textbf{a-b}. Automated evaluation results based on two SOTA LLMs. 
    Proportion of valid triples (score$\geq$2.0) across relation types as assessed by GPT-4.1 (a) and DeepSeek-v3 (b) on a randomly selected subset of 34,725 abstracts (83,438 triples). 
    \textbf{c-d}. 
    Manual evaluation by two independent domain experts. 
    Proportion of valid triples across relation types as assessed by Expert 1 (c) and Expert 2 (d) on a subset of 1,000 abstracts (2,767 triples).
    \textbf{e-f}. 
    Performance of GPT-4.1 and DeepSeek-v3 compared to two expert evaluations on the shared evaluation subset, reporting precision, recall, and F1 score. 
    \textbf{g}. Pairwise inter-rater agreement between experts and LLMs quantified by Cohen's kappa coefficients, demonstrating substantial consistency across all evaluators.}
    \label{fig:evaluate}
\end{figure}

For automated evaluation, two SOTA LLMs, GPT-4.1 \cite{openai_gpt4.1} and DeepSeek-v3 \cite{liu2024deepseek}, were employed. 
A random subset comprising 1\% of the abstracts (n = 34,725), resulting in 83,438 extracted triples, was selected for evaluation. 
Each abstract and its corresponding triples were formatted into structured prompts and independently assessed by both models according to a standardized four-tier rubric: {Correct (3.0)}, {Likely Correct (2.0)}, {Likely Incorrect (1.0)}, and {Incorrect (0.0)} (the specific evaluation prompt is illustrated in Supplementary Figure 7 \textbf{a}). 
Triples receiving scores of $\geq 2.0$ were deemed valid. 
The evaluation outcomes are presented in Figure \ref{fig:evaluate} \textbf{a} and \textbf{b}, illustrating the proportion of valid triples across relation types for GPT-4.1 and DeepSeek-v3, respectively. 
Both models demonstrated high overall accuracy, with 85.44\% and 88.10\% of triples rated as valid by GPT-4.1 and DeepSeek-v3, respectively. 
For most relation types, validity was approximately 90\%, except for Negative\_Correlate, which exhibited slightly lower agreement. 
These findings underscore the high precision of the Extractor Agent across diverse biomedical relation types and support its utility for downstream analyses.

In parallel, a manual evaluation was performed to further assess extraction accuracy. 
Two PhD-level annotators with interdisciplinary expertise in medicine and computer science independently reviewed a subset of 1,000 abstracts, encompassing 2,767 extracted triples. 
Each abstract and its associated triples were assessed using the same four-tier scoring rubric, with triples receiving scores $\geq$2.0 deemed valid. The evaluation results are shown in Figure \ref{fig:evaluate} \textbf{c}, \textbf{d}, which report the proportion of valid triples across relation types for Expert 1 and Expert 2, respectively. 
The two reviewers exhibited high concordance, with overall validity rates exceeding 86\% for both assessors. 
For most relation types, validity rates were around 90\%. 
The strong agreement between manual and automated evaluations further supports the robustness of the Extractor Agent in accurately capturing biomedical relationships, underscoring the reliability of the extracted knowledge for large-scale medical analyses.

To further validate the reliability of the LLM-based assessments, we used two expert annotations as reference standards to evaluate GPT-4.1 and DeepSeek-v3 on the same subset of 1,000 abstracts, respectively. 
As shown in Figure \ref{fig:evaluate} \textbf{e}, \textbf{f}, both models exhibited strong concordance with expert evaluations, achieving precision, recall, and F1 scores of approximately 95\% across metrics. 
These results further corroborate the accuracy of the automated scoring framework and its alignment with expert judgment.

Finally, inter-rater agreement across all evaluators---including two human experts and two LLMs---was quantified using pairwise Cohen's kappa coefficients computed on a shared evaluation subset (Figure \ref{fig:evaluate} \textbf{g}) \cite{mchugh2012interrater}. 
Most pairwise comparisons yielded kappa values above 0.6, corresponding to substantial agreement, a commonly accepted benchmark for reliable concordance in domains involving subjective judgment, including medicine, psychology, and natural language processing \cite{landis1977measurement}. The remaining comparisons fell marginally below this threshold but remained within the moderate agreement range and close to the substantial category. Together, these findings indicate strong inter-rater reliability across both human and automated evaluators, supporting the robustness and reproducibility of the evaluation framework.

\subheading{Evaluating Downstream Utility in Medical Question Answering}

\begin{figure}[h]
    \centering
    \includegraphics[width=1.0\textwidth]{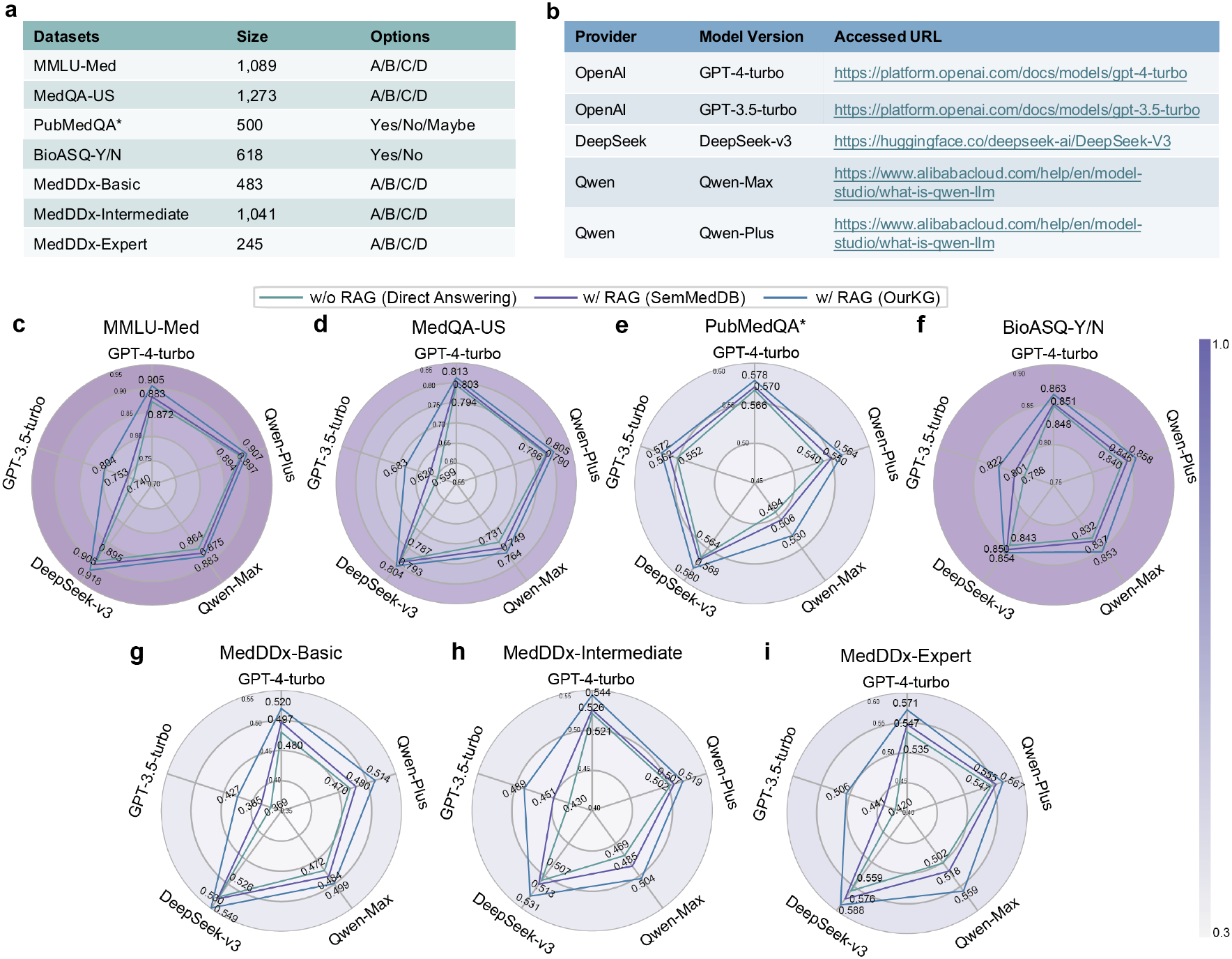} 
    \caption{Evaluation of knowledge-augmented QA across seven medical QA benchmarks. 
    \textbf{a}. Dataset statistics, including the number of questions in each benchmark. 
    \textbf{b.} Summary of evaluated LLMs and access endpoints. 
    \textbf{c–f}. Radar plots showing accuracy on MMLU-Med, MedQA-US, PubMedQA*, and BioASQ-Y/N. 
    \textbf{g–i}. Radar plots showing accuracy on MedDDx-Basic, MedDDx-Intermediate, and MedDDx-Expert. 
    For each dataset, results are reported under three settings: direct answering without retrieval, RAG using SemMedDB as the retrieval source, and RAG using OurKG constructed by the MedKGent framework as the retrieval source. 
    Values indicate accuracy (correct answers divided by total questions). 
    Statistical significance is assessed using paired bootstrap resampling (5,000 iterations) over QA instances; 
    improvements are considered significant when p$<$0.05, where applicable.}
    \label{fig:rag_qa}
\end{figure}

We evaluated the downstream utility of our KG as a RAG information source across seven multiple-choice medical QA datasets. 
These include four widely used datasets \cite{xiong2024benchmarking}---MMLU-Med, MedQA-US, PubMedQA*, and BioASQ-Y/N---covering a broad range of medical knowledge and reasoning tasks, as well as MedDDx, a newly introduced diagnostic reasoning benchmark suite \cite{su2025kgarevion}. 
MedDDx is stratified into three subsets (Basic, Intermediate, and Expert) according to the semantic similarity among answer options, with higher levels reflecting increased diagnostic ambiguity and reduced risk of training data leakage. 
Dataset statistics are summarized in Figure \ref{fig:rag_qa} \textbf{a}. 
We evaluated five leading LLMs---GPT-4-turbo, GPT-3.5-turbo \cite{openai_gpt_turbo}, DeepSeek-v3 \cite{liu2024deepseek}, Qwen-Max, and Qwen-Plus \cite{yang2025qwen3}---under three settings: 
(1) direct answering without retrieval, 
(2) RAG using SemMedDB \cite{kilicoglu2012semmeddb} as the retrieval source, 
and (3) RAG using our KG as the retrieval source. 
All models were tested in a zero-shot setting via publicly available APIs without additional fine-tuning, and identical retriever configurations and prompts were used across all conditions. 
Version details and access endpoints are summarized in Figure \ref{fig:rag_qa} \textbf{b}.

Figures \ref{fig:rag_qa} \textbf{c–i} present radar plots comparing QA accuracy (the proportion of correctly answered questions) across datasets. 
Across all benchmarks and models, both knowledge-augmented settings outperform direct answering, with RAG using OurKG constructed by the MedKGent framework consistently achieving the strongest performance. 
SemMedDB-based RAG provides modest but consistent improvements over direct answering, whereas MedKGent yields larger gains, particularly on tasks requiring richer biomedical and diagnostic knowledge. 
On MedQA-US, OurKG-based RAG improves accuracy by up to 8.4 percentage points over direct answering (\emph{e.g.}, GPT-3.5-turbo: 0.599 to 0.683) and consistently outperforms SemMedDB-based RAG across all evaluated models. 
Similar trends are observed on the MedDDx suite, where OurKG achieves the largest improvements on the Expert subset, highlighting its effectiveness in supporting complex differential diagnosis. 
As reflected by the lighter background shading in Figures \ref{fig:rag_qa} \textbf{g–i} compared with Figures \ref{fig:rag_qa}\textbf{c–f}, all models exhibit lower accuracy on MedDDx than on the widely used benchmarks, consistent with the higher semantic complexity and reduced risk of training data leakage inherent to the MedDDx design. 
Notably, several models perform better on MedDDx-Expert than on MedDDx-Basic, despite the former involving greater semantic ambiguity. 
This counterintuitive pattern may be attributable to differences in distractor construction, where Expert-level distractors, although lexically similar, exhibit greater internal consistency that better aligns with the models' reasoning strategies. 
On knowledge-intensive but less ambiguous tasks such as MMLU-Med and BioASQ-Y/N, improvements are smaller but remain consistent, with OurKG achieving the highest overall accuracy among all evaluated settings. 

Overall, these results indicate that while an established biomedical KG such as SemMedDB can provide useful contextual support for QA, the temporally evolving and confidence-aware knowledge integrated by the MedKGent framework offers greater downstream effectiveness for medical reasoning tasks.

\section{Discussion} \label{discussion}

The rapid growth and increasing complexity of biomedical literature pose persistent challenges for organizing, synthesizing, and extracting actionable knowledge from large volumes of unstructured text. 
MedKGent addresses these challenges by introducing an LLM-agent framework for constructing a temporally evolving medical KG that preserves publication-level provenance and accumulates evidence over time. 
The principal contributions of this work are methodological and infrastructural: 
(i) a two-agent architecture that decouples entity detection and normalization from LLM-based relation inference and incremental graph construction; 
(ii) a sampling-based confidence estimation and evidence accumulation mechanism that reinforces recurring findings and resolves relational conflicts in a temporally informed manner; 
and (iii) a systematic evaluation pipeline that combines automated LLM-based assessment with independent manual review to validate extraction quality at scale.

We intentionally use PubMed abstracts as the primary data source. 
Abstracts are concise, information-dense, and uniformly available with structured metadata, which together enable large-scale, reproducible KG construction spanning multiple decades of biomedical research. 
This abstract-first design substantially improves scalability while reducing practical barriers associated with full-text acquisition, including licensing restrictions, document heterogeneity, and parsing complexity. 
At the same time, it introduces a clear limitation: the constructed KG reflects information available in abstracts, but not necessarily details contained in full-text bodies, supplementary materials, or figures and tables. 
Accordingly, this manuscript is positioned as a methods and infrastructure contribution. 
The resulting KG serves as a high-quality, temporally aware literature resource that supports literature-grounded downstream tasks (\emph{e.g.}, knowledge-augmented medical QA and hypothesis generation), but it is not intended as a substitute for direct clinical evidence or prospective clinical validation.

Automated and expert evaluations demonstrate that MedKGent produces high-quality relational triples, with validity rates approaching 90\% across relation types, and that the resulting KG consistently improves performance in RAG across multiple medical QA benchmarks. 
These findings support the suitability of the KG as an infrastructure component for literature-centered AI applications. 
At the same time, the comparison with SemMedDB primarily evaluates downstream retrieval utility rather than providing a comprehensive graph-level assessment of KG construction quality. 
A full KG-to-KG comparison would require constructing alternative large-scale biomedical KGs from the same corpus of over 10 million PubMed abstracts using other modern pipelines, followed by systematic comparisons of entity coverage, relation quality, graph structure, and downstream performance. 
Such benchmarking would involve substantial computational resources and engineering effort for entity normalization, relation extraction, and conflict resolution at a scale comparable to MedKGent. 
Therefore, retrieval-based evaluation should be interpreted as a scalable and application-oriented assessment of KG utility, not as a complete proxy for KG construction quality. 
These findings also motivate subsequent application studies, which must be designed and evaluated under domain-appropriate and task-specific protocols.

We explicitly discuss several methodological trade-offs and limitations that informed our design choices. 
With respect to abstract-only versus full-text extraction, full texts and supplementary materials often contain richer and more detailed evidence, including dosing information, subgroup analyses, methodological details, and structured data embedded in tables and figures, which can be critical for mechanistic inference and certain downstream tasks, particularly fine-grained drug-repurposing analyses. 
However, large-scale full-text ingestion presents substantial practical challenges, including legal and licensing constraints, heterogeneous document structures, the need for table and figure parsing (often involving OCR), increased noise and redundancy, and significantly higher computational and storage costs. 
We therefore adopt an abstracts-first strategy as a pragmatic and reproducible baseline, and plan PubMed Central (PMC) full-text integration as a staged extension requiring specialized parsers and provenance-aware evidence aggregation.

Regarding entity detection and normalization, we rely on PubTator3 for initial entity detection and normalization to stable identifiers (\emph{e.g.}, MeSH and NCBI), simplifying downstream entity linking and retrieval. 
This decoupled design improves robustness and efficiency, but may propagate upstream tool biases and fail to capture complex or context-dependent entity mentions. 
Joint-extraction approaches or LLM-augmented entity refinement represent promising alternatives and are important directions for future investigation.

With respect to temporal leakage and evaluation design, time-aware KG construction reduces the risk of temporal leakage but does not eliminate it entirely. 
Any retrospective predictive evaluation, such as literature-based drug-repurposing, must enforce strict publication cut-off protocols, exclude future evidence during candidate generation, and construct validation sets from subsequent literature or clinical-trial registries using conservative criteria. 
We outline a minimal experimental checklist below to guide the design of such studies.

LLM-specific reliability concerns, including hallucinated or weakly supported relations, must also be considered. 
LLMs may generate plausible yet unsupported relations, particularly in low-evidence or ambiguous contexts. 
We mitigate this risk through sampling-based consistency checks, structured prompting, and targeted manual review. 
In addition, our quality assessment partly relies on LLM-assisted evaluation, which should be interpreted as a scalable but imperfect strategy for large-scale KG assessment. 
Exhaustive expert review of nearly three million triples is infeasible, but we reduce the risk of self-confirming evaluation by using evaluator models distinct from the Qwen2.5-32B model used for KG construction and by validating LLM-based judgments against an expert-annotated subset. Nonetheless, further work on adversarial testing, model calibration, cross-model consensus, and external expert-annotated benchmarks remains necessary to improve reliability.

These methodological considerations also have direct implications for translating the constructed KG into real-world clinical settings. 
Translating MedKGent into direct clinical decision support requires efforts that extend beyond literature extraction. 
A clinically oriented evaluation or deployment would need to include precise alignment between KG concepts and clinical terminologies used in Electronic Health Record (EHR) systems (\emph{e.g.}, SNOMED CT, RxNorm, and LOINC), privacy-preserving pipelines or federated evaluation mechanisms for interacting with EHR data, selection of clinically meaningful endpoints such as diagnostic accuracy, decision concordance, or patient-relevant outcomes, and clinician-led adjudication combined with prospective or rigorously designed retrospective validation. 
A practical near-term pathway is to evaluate the KG as an augmenting knowledge source for clinically oriented QA research systems, offline decision-support safety checks, or cohort-discovery research tools.
Demonstrating direct impact on clinical outcomes, however, will require multidisciplinary collaboration, institutional partnerships, and appropriate governance frameworks.

To conduct robust and reproducible quantitative literature-based drug-repurposing analyses, we recommend explicit task framing (drug-centric versus disease-centric), strict temporal holdouts with clearly defined prediction cut-off dates, construction of validation sets from subsequent literature and clinical-trial registries curated under conservative inclusion rules, quantitative evaluation metrics beyond simple counts (including recall@k, observed positive rate, time-to-validation distributions, and precision at different operating points), and transparent reporting of false positives and candidate prioritization criteria. 
While integrating PMC full text and additional biomedical databases is likely to improve sensitivity for mechanistic signals, doing so requires substantial engineering and curation effort and is therefore planned as a separate, focused study.

Building on the current KG, we plan three major extensions: PMC full-text ingestion, including the development of robust full-text parsers (HTML/PDF), section-aware extraction, and table/figure parsing with provenance scoring; EHR and clinical-terminology alignment, involving mapping KG concepts to clinical terminologies and exploring privacy-preserving EHR linkage or federated evaluation strategies; and rigorous application studies, comprising dedicated quantitative drug-repurposing experiments and pilot clinically oriented, offline use-case evaluations in collaboration with clinical partners. 
Each extension will require additional dataset engineering, carefully designed evaluation protocols, and appropriate governance.

In summary, MedKGent provides a scalable and temporally informed framework for converting large-scale biomedical abstracts into a high-quality medical KG that improves literature-grounded AI tasks. 
The system is intentionally positioned as infrastructure: it enables downstream applications but does not itself constitute clinical validation. 
Extending the framework to richer evidence sources, such as full text and EHRs, and conducting rigorous, application-specific evaluations represent important next steps that will be pursued in follow-up work.

\section{Methods}\label{method}

\subheading{Data Preprocessing}

We selected PubMed abstracts as the primary data source for constructing the medical KG, owing to their concise yet information-dense summaries of research findings. 
The structured format and compact content of these abstracts make them particularly suitable for large-scale knowledge extraction. 
Over 20 million abstracts were collected from the \href{https://pubmed.ncbi.nlm.nih.gov/}{PubMed}, which integrates biomedical literature with specific publication dates. After excluding entries that failed to download or lacked abstract content---commonly due to missing metadata in older records---we retained more than 16 million abstracts for further analysis.

We first analyzed the length distribution of the abstracts (Supplementary Figure 5 \textbf{a}), which followed a characteristic long-tail pattern. 
A small proportion of abstracts exceeded 500 words; for clarity, all abstracts longer than 500 words were aggregated into a single group, while a considerable number were shorter than 100 words. 
Manual inspection indicated that both extremes often contained irregular or uninformative content. 
To improve consistency and reduce noise, we retained only abstracts within the 100–300 word range, the most common length observed. 
Next, We analyzed the distribution of publication years (Supplementary Figure 5 \textbf{b}). 
Articles published before 1965 were rare and were aggregated for reference. 
A notable increase in publication volume began in 1975, while records from 2024 were incomplete at the time of data collection. 
As a result, we restricted our dataset to abstracts published between 1975 and 2023. 
Following a series of quality control procedures---including length filtering, temporal constraints, and content-based screening---we retained a total of 10,014,314 PubMed abstracts. 
These abstracts were organized into a fine-grained daily time series from January 1, 1975, to December 31, 2023 (Supplementary Figure 5 \textbf{c}), facilitating high-resolution temporal analysis of the emergence of medical knowledge. 
To ensure intra-day consistency, abstracts published on the same date were sorted by ascending PubMed ID.

\subheading{MedKGent Framework}

We developed MedKGent, an LLM-based agent framework for constructing a temporally evolving medical KG. 
MedKGent processes biomedical abstracts sequentially from January 1, 1975, to December 31, 2023, allowing the KG to grow incrementally in step with the historical development of medical knowledge while remaining extensible to future updates. 
The framework comprises two coordinated agents---the Extractor Agent and the Constructor Agent---deployed via a self-hosted API based on the open-source Qwen2.5-32B-Instruct model \cite{qwen2.5}. 
Deployment leveraged 48 NVIDIA L20Z GPUs (80 GB each), with two services per GPU, totaling 96 services. 
This setup significantly reduces costs compared to commercial LLM APIs, particularly given the large volume of abstracts processed, and provides greater flexibility for system optimization, including improved message queue throughput and processing speed. 
Collectively, the two agents extract structured medical facts and integrate them into a dynamic, time-aware KG.

\subheading{Extractor Agent}

The Extractor Agent identifies biomedical entities within each abstract using PubTator3 \cite{wei2024pubtator}, an off-the-shelf AI tool developed and continuously maintained by the NCBI. 
This module annotates six categories of biomedical entities---genes, diseases, chemicals, variants, species, and cell lines---producing an entity set $E=\{e_1, e_2, \cdots\}$ extracted from a given abstract $A$. 
While LLMs can jointly extract entities and relational triples via prompting, the Extractor Agent adopts a decoupled strategy: entities are first annotated using the PubTator3 tool, followed by relation extraction with the LLM. 
This design offers several key advantages. 
First, PubTator3 not only detects biomedical concepts but also assigns unique identifiers, normalizing synonymous mentions to standardized terminology entries. 
These identifiers facilitate downstream tasks such as entity disambiguation, merging, and retrieval---critical steps for accurate graph construction by the Constructor Agent. 
The terminologies used for each entity type are summarized in Supplementary Figure 6 \textbf{b}. 
Second, PubTator3 provides SOTA entity annotations across PubMed abstracts and is widely adopted within the biomedical research community. 
Third, decoupling the tasks allows the LLM to focus exclusively on relation extraction, reducing cognitive load and enhancing overall performance relative to joint extraction approaches.

Given both the abstract $A$ and its extracted entity set $E$, the Extractor Agent then prompts the LLM to infer semantic relationships between entity pairs, guided by a predefined set of relation types $R$ and their textual definitions. 
Leveraging the LLM's internal knowledge, it assigns a relation $r_n\in R $ to each relevant entity pair $(e_i,e_j)\in E$, resulting in a set of candidate relational triples $T_{\text{candi}}=\{t_1, t_2, \cdots | t_k = (e_i, r_n, e_j)\}$. 
The prompting template used for this inference is illustrated in Supplementary Figure 6 \textbf{a}. 
We define a set of 12 core biomedical relation types, comprising seven bidirectional relations---Associate, Negative\_Correlate, Positive\_Correlate, Compare, Cotreat, Interact, and Drug\_Interact---and five unidirectional relations---Cause, Inhibit, Treat, Stimulate, and Prevent. 
The properties and descriptions of these relations are detailed in Supplementary Figure 6 \textbf{d}. 
These relation types can be flexibly extended through prompt design as needed. 
This adaptable design eliminates the need for rigid schema definitions or retraining, as required in traditional supervised pipelines, allowing MedKGent to incorporate novel and evolving medical relations with minimal manual intervention.

To assign an initial confidence score to each extracted relational triple, the Extractor Agent employs a sampling-based confidence estimation strategy during LLM inference, inspired by the self-consistency principle \cite{wangself,taubenfeld2025confidence,chen2024universal,li2025system}. 
Specifically, for an extraction prompt, the agent performs $N$ parallel stochastic inferences (with $N$=$50$), yielding $N$ sets of candidate triples, denoted as $\{T^1_{\text{candi}}, \cdots, T^N_{\text{candi}}\}$. 
The underlying assumption is that relational triples consistently generated across multiple runs reflect higher model certainty, whereas spurious or unstable outputs occur less frequently. 
To encourage output diversity, the LLM within the Extractor Agent operates with a moderate temperature coefficient ($\tau$=$0.7$), increasing output variability across repeated inferences. 
This diversity enables more discriminative frequency-based confidence estimation, as triples that recur despite stochastic variation are more likely to be reliable. 
Following this, the Extractor Agent conducts a formatting check on all candidate triples, eliminating those containing extraneous or irrelevant characters. 
For the remaining triples, the agent computes initial confidence scores based on their frequency of occurrence across the $N$ inference runs. 
Each frequency is normalized by mapping it to the nearest lower multiple of $0.05$, which is then assigned as the triple's initial confidence score. 
Triples with confidence scores below $0.6$ are discarded, ensuring that only relations supported by a clear majority of inference runs are retained for downstream graph construction.

For each retained triple, the agent enriches both head and tail entities with two key attributes, Exact Keywords and Semantic Embedding, in addition to the original name, entity type, unique identifier, and terminology. 
Exact Keywords list all textual variants of the entity within the abstract (standardized to lowercase and mapped to a single identifier), while Semantic Embedding represents the entity as a 768-dimensional vector generated using the BiomedNLP-BiomedBERT-base-uncased-abstract-fulltext model \cite{pubmedbert}. 
These enriched representations improve the precision and efficiency of biomedical information retrieval, particularly when explicit entity identifiers are unavailable. 
Each triple is further annotated with the PubMed ID of the source abstract and a timestamp corresponding to the publication date, ensuring full traceability and source attribution. 
The final set of enriched triples is then passed to the Constructor Agent for integration into the KG.

\subheading{Constructor Agent}

The Constructor Agent incrementally integrates relational triples extracted by the Extractor Agent into a dynamically evolving temporal KG through continuous interaction with a Neo4j graph database. 

For each relational triple, the Constructor Agent first checks whether the head and tail entities---identified by their unique identifiers---already exist in the graph. 
If either entity is absent, the corresponding node is inserted, and a new edge is created to represent the specified relationship. 
If both entities are present and the relation type matches that of an existing edge, the triple is interpreted as a recurrence of the same medical knowledge in a subsequent publication. 
In this case, the graph is updated by increasing the edge's confidence score $s$ using the following enhancement function:
\begin{equation}
    s = 1 - (1-s)*(1-s') \text{,}
    \label{eq:update_conf}
\end{equation}
where $s'$ is the confidence score of the newly observed triple. 
This formulation ensures that confidence increases monotonically as evidence accumulates; the more frequently the same knowledge appears during graph construction, the higher its resulting confidence. 
The PubMed ID associated with the new occurrence is appended to the edge's PubMed ID list of supporting references, and the Timestamp is updated to reflect the most recent publication. 
Historical provenance is preserved through the complete PubMed ID list, enabling full traceability of supporting evidence over time. 
Conversely, if both entities exist but the relation type differs from that of the existing edge, this suggests that multiple relationships have been assigned to the same entity pair. 
To avoid redundancy and inconsistency, we assume that only one, most appropriate relation should be maintained. 
In such cases, the Constructor Agent invokes an LLM to resolve the conflict by selecting the more suitable relation, considering the confidence score $s$ and timestamp $t$ of the existing edge, as well as the confidence score $s'$ and timestamp $t'$ of the incoming triple. 
To ensure more deterministic output than that used in the Extractor Agent, the LLM in the Constructor Agent is configured with a lower temperature parameter ($\tau=0.2$). 
The prompting strategy for this decision process is illustrated in Supplementary Figure 6 \textbf{c}.

\subheading{Quality Assessment}

We assessed the quality of relational triples extracted by the Extractor Agent through both automated and manual evaluations, leveraging two SOTA LLMs---GPT-4.1 \cite{openai_gpt4.1} and DeepSeek-v3 \cite{liu2024deepseek}---as well as two PhD-level annotators with interdisciplinary expertise in medicine and computer science. 
For each medical abstract and its corresponding set of extracted triples, individual triples were evaluated using a standardized four-level scoring rubric: 3.0 (Correct), 2.0 (Likely Correct), 1.0 (Likely Incorrect), and 0.0 (Incorrect). 
The evaluation prompt provided to both LLMs and human annotators is illustrated in Supplementary Figure 7 \textbf{a}. 

A relational triple was defined as \textit{valid} if it received a score of $\geq 2.0$. 
The validity rate was calculated as:
\begin{equation}
    Validity\ Rate = \frac{Number\ of\ triples\ with\ score\geq 2.0}{Total\ number\ of\ evaluated\ triples} \text{.}
\end{equation} 
To assess the reliability of automatic evaluation, we compared LLM-based assessments with human annotations on a shared evaluation subset, treating human judgments as ground truth. 
The precision, recall, and $F_1$-score of the automatic evaluations were computed as:
\begin{equation}
    Precision = \frac{TP}{TP+FP},\ Recall=\frac{TP}{TP+FN},\ F_1 = \frac{2\times Precision\times Recall}{Precision+Recall}\text{,} 
\end{equation}
where TP, FP, and FN represent true positives, false positives, and false negatives, respectively. 
To further quantify inter-rater agreement, we calculated Cohen's kappa coefficient \cite{mchugh2012interrater} for each pair of evaluators, including both LLMs and human annotators, resulting in six pairwise comparisons across the four raters. 
The Kappa coefficient was computed as:
\begin{equation}
   \kappa = \frac{p_0-p_e}{1-p_e}\text{,} 
\end{equation}
where $p_0$ represents the observed agreement and $p_e$ denotes the expected agreement by chance. This analysis provides a quantitative measure of rating consistency across evaluators.

\subheading{Retrieval-Augmented Generation}

The constructed KG serves as a reliable external source for information retrieval and can be integrated into LLMs via a RAG framework. 
By providing structured biomedical context, the KG enhances LLM performance across a range of medical QA benchmarks. 

Given a user query $q$, we first extract the set of medical entities present in the question, denoted as $E^q = \{e^q_1, e^q_2, \cdots\}$. 
When using PubTator3 \cite{wei2024pubtator}---the same entity recognition tool employed during KG construction---each extracted entity is assigned a unique identifier. 
This allows for efficient entity linking by matching these identifiers to the corresponding nodes $N^q = \{n^q_1, n^q_2, \cdots\}$ within the graph. 
Alternatively, if medical entities are extracted using other methods---such as prompting an LLM---they may lack standardized identifiers. 
In such cases, the extracted entity mentions are first converted to lowercase and matched against the Exact Keywords attribute of each node in the KG. 
A successful match enables linkage of the entity to the corresponding graph node. 
In both approaches, if an entity cannot be linked via its identifier or if its surface form does not appear in any node's Exact Keywords list, we apply a semantic similarity strategy to complete the entity linking process. 
Specifically, the embedding of the query entity is computed using the same model employed for generating node-level semantic representations (\emph{i.e.}, BiomedNLP-BiomedBERT-base-uncased-abstract-fulltext \cite{pubmedbert}) and is compared against the Semantic Embedding of all nodes in the KG using cosine similarity. 
The entity is then linked to the node with the highest cosine similarity score, which may correspond to either the exact concept or a semantically related medical entity.
This entity linking framework---combining identifier-based matching, lexical normalization, and semantic embedding---ensures robust and flexible integration of KG-derived knowledge into downstream QA tasks.

Following entity linking, we construct evidence subgraphs using a neighbor-based exploration strategy \cite{wen2024mindmap} to enhance the reasoning capabilities of LLMs. 
For each entity-linked node in the query-specific set $N^q$, we retrieve its one-hop neighbors within the KG. 
Specifically, for each node $n^q_i\in N^q$, all adjacent nodes $n^{q'}_i$ are identified, and the corresponding triples $(n^q_i,r,n^{q'}_i)$ are appended to form a localized subgraph $G^q_i$. 
This expansion captures the immediate relational context surrounding the query entities, which is essential for enabling fine-grained medical reasoning. 
The complete evidence set for a given query is then defined as the union of these localized subgraphs: $G^q=\{G^q_1,G^q_2,\cdots\}$. 
The resulting subgraph $G^q$ may contain a large number of relational triples, including redundant or irrelevant information, which can adversely impact LLM reasoning \cite{wueasily}. 
To address this, we leverage the LLM's inherent ranking capability to selectively filter high-value knowledge \cite{sun2023chatgpt}. 
Given the question $q$ and its corresponding subgraph $G^q$, we prompt the LLM to rerank the triples based on their relevance to the query. 
Only the top k triples ($k=5$) are retained to form a refined subgraph $G^q_{rerank}$. 
The reranking prompt is shown in Supplementary Figure 7 \textbf{b}. 
In the final reasoning stage, the question $q$ and the refined subgraph $G^q_{rerank}$ are provided to the LLM for answer generation. 
This targeted context significantly reduces cognitive noise and improves answer quality. 
The final inference prompt is illustrated in Supplementary Figure 7 \textbf{c}.

\subheading{Ethics statement}

This study used only publicly available bibliographic data (PubMed abstracts) and did not involve human participants, patient-level clinical data, or identifiable personal information. Accordingly, ethics approval and informed consent were not required.

\section*{Data availability}
The medical knowledge graph constructed in this study is publicly available at \url{https://huggingface.co/datasets/ShowerMaker/MedKGent-KG}.

\section*{Code availability}

The source code for MedKGent is available at \url{https://github.com/BladeDancer957/MedKGent}. An archived, citable release is available at [10.5281/zenodo.20759287].

\section*{Acknowledgements} 
The authors received no specific funding for this work.

\section*{Author contributions}

D.Z. conceived the method, collected and pre-processed the data, implemented the core MedKGent framework, constructed the KG, and wrote the manuscript. 
Z.W. performed automatic quality assessment of the extracted relational triples, evaluated downstream utility in medical QA, and contributed to the analysis and interpretation of downstream evaluation results. 
Z.L. was responsible for deploying the LLM API and optimizing the efficiency of relational triple extraction. 
Y.Y. and S.J. contributed to the preparation of figures and data visualization. 
H.X. and X.W. provided the GPU resources for API deployment and offered guidance on optimization and acceleration. 
J.D., Y.Z., T.Z. and J.Y. served as intermittent advisors, offering suggestions on experimental design and manuscript writing.
X.C. and L.S. designed the overall study, served as corresponding authors, and played a key role in coordinating and integrating all resources. 
All the authors read and approved the manuscript.

\section*{Competing interests}
Jie Yang is a Guest Editor of the collection to which this manuscript was submitted and was not involved in the journal's review of, or decisions related to, this manuscript. 
The other authors declare no competing financial or non-financial interests.

\bibliographystyle{unsrt} 
\bibliography{reference}

\newpage

\newpage

\begin{extendedfigure*}
    \centering
    \includegraphics[width=1\linewidth]{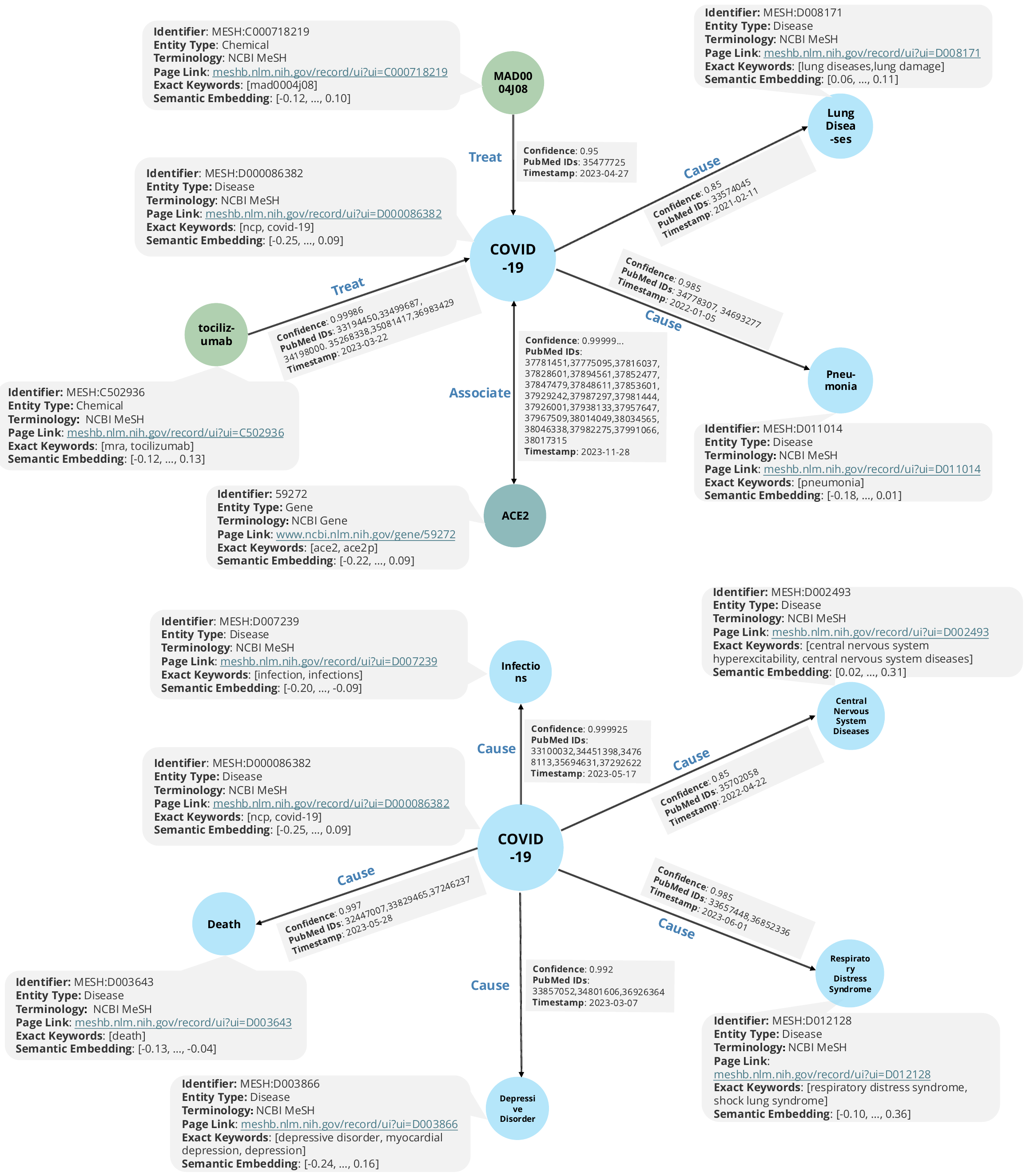}
    \caption{Local subgraph visualization centered on COVID-19, displaying surrounding relationship triples. Each node is characterized by key attributes, including Identifier, Entity Type, Terminology, Page Link, Exact Keywords, and Semantic Embedding, facilitating efficient entity linking through exact or similarity matching. The relationships in the KG are further enriched by attributes such as Confidence, PubMed IDs, and Timestamp, enhancing traceability, accuracy, and temporal relevance.}
\end{extendedfigure*}

\begin{extendedfigure*}
    \centering
    \includegraphics[width=1\linewidth]{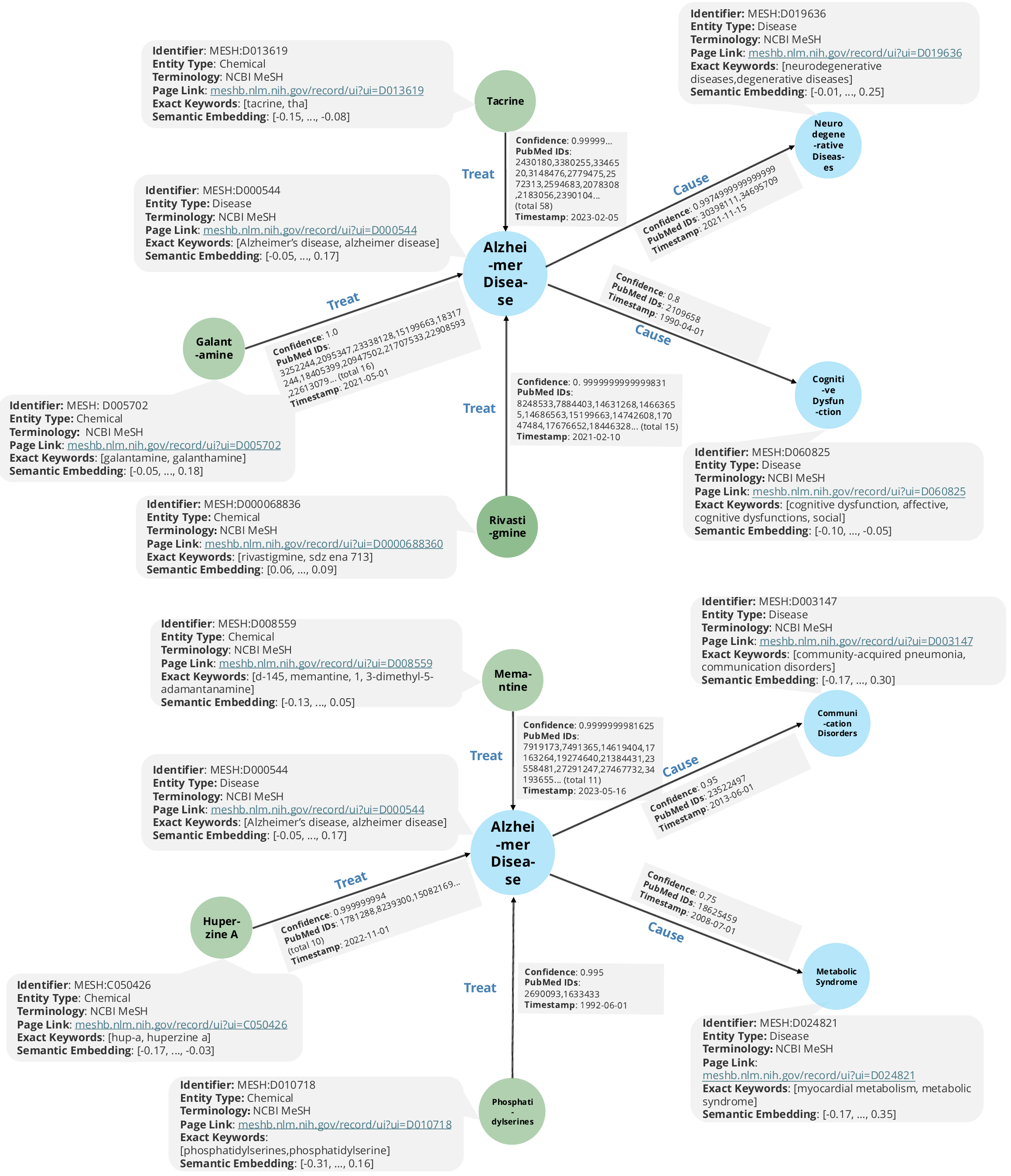}
    \caption{Local subgraph visualization centered on Alzheimer's Disease, displaying surrounding relationship triples. Each node is characterized by key attributes, including Identifier, Entity Type, Terminology, Page Link, Exact Keywords, and Semantic Embedding, facilitating efficient entity linking through exact or similarity matching. The relationships in the KG are further enriched by attributes such as Confidence, PubMed IDs, and Timestamp, enhancing traceability, accuracy, and temporal relevance.}
\end{extendedfigure*}

\begin{extendedfigure*}
    \centering
    \includegraphics[width=1\linewidth]{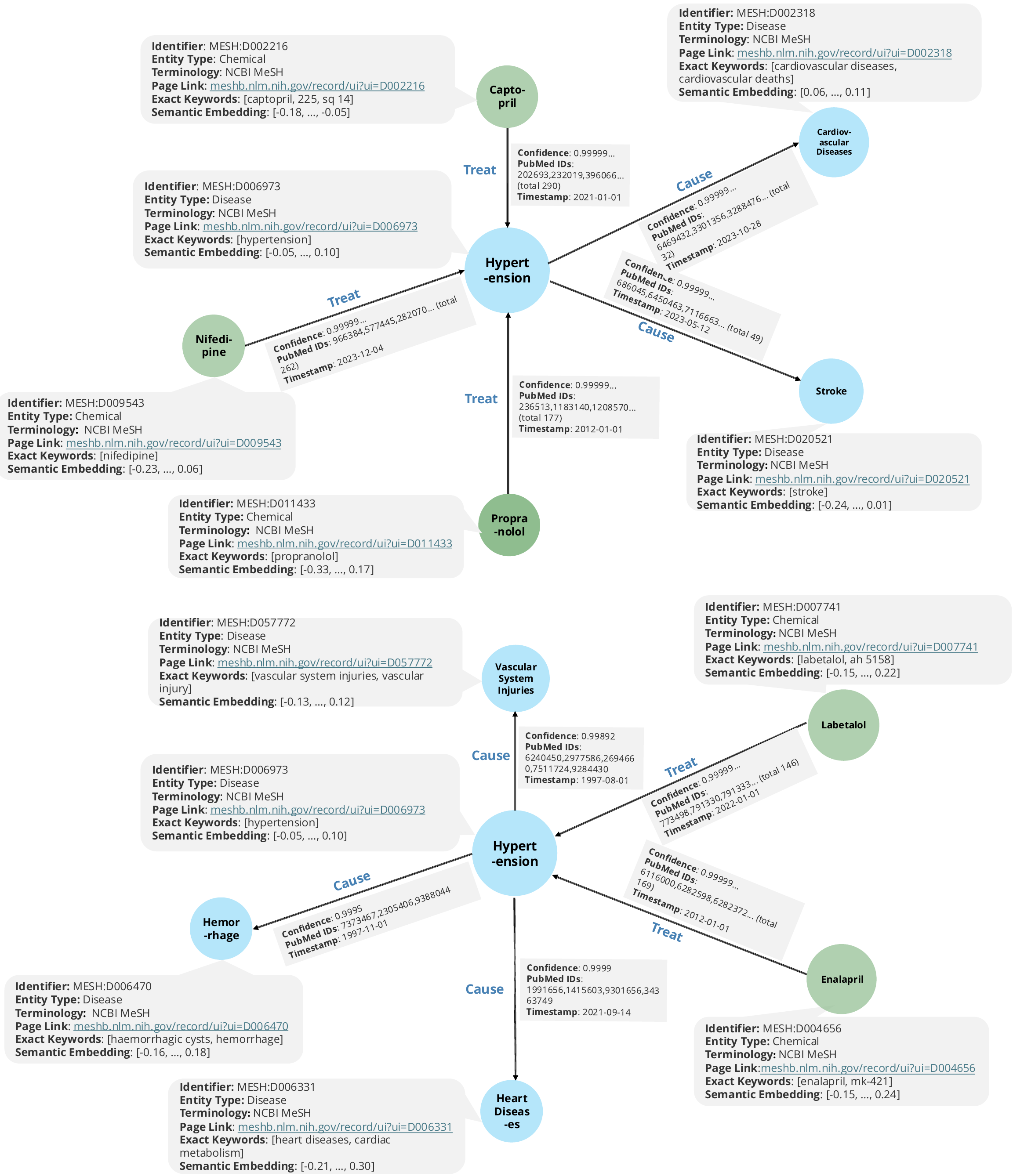}
    \caption{Local subgraph visualization centered on Hypertension, displaying surrounding relationship triples. Each node is characterized by key attributes, including Identifier, Entity Type, Terminology, Page Link, Exact Keywords, and Semantic Embedding, facilitating efficient entity linking through exact or similarity matching. The relationships in the KG are further enriched by attributes such as Confidence, PubMed IDs, and Timestamp, enhancing traceability, accuracy, and temporal relevance.}
   
\end{extendedfigure*}

\begin{extendedfigure*}
    \centering
    \includegraphics[width=1\linewidth]{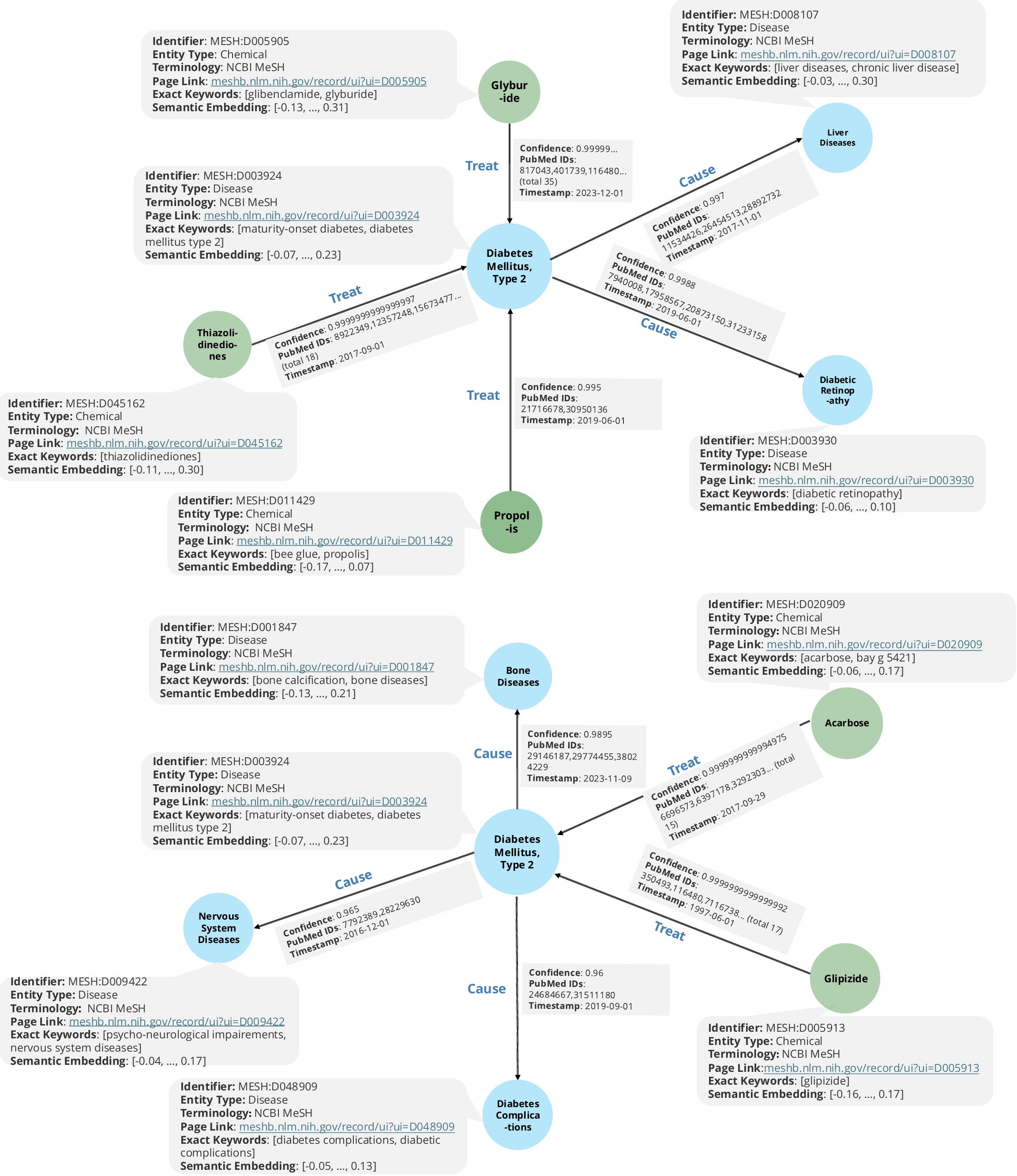}
    \caption{Local subgraph visualization centered on Diabetes Mellitus, displaying surrounding relationship triples. Each node is characterized by key attributes, including Identifier, Entity Type, Terminology, Page Link, Exact Keywords, and Semantic Embedding, facilitating efficient entity linking through exact or similarity matching. The relationships in the KG are further enriched by attributes such as Confidence, PubMed IDs, and Timestamp, enhancing traceability, accuracy, and temporal relevance.}
 
\end{extendedfigure*}

\begin{extendedfigure}
    \centering
    \includegraphics[width=1.0\textwidth]{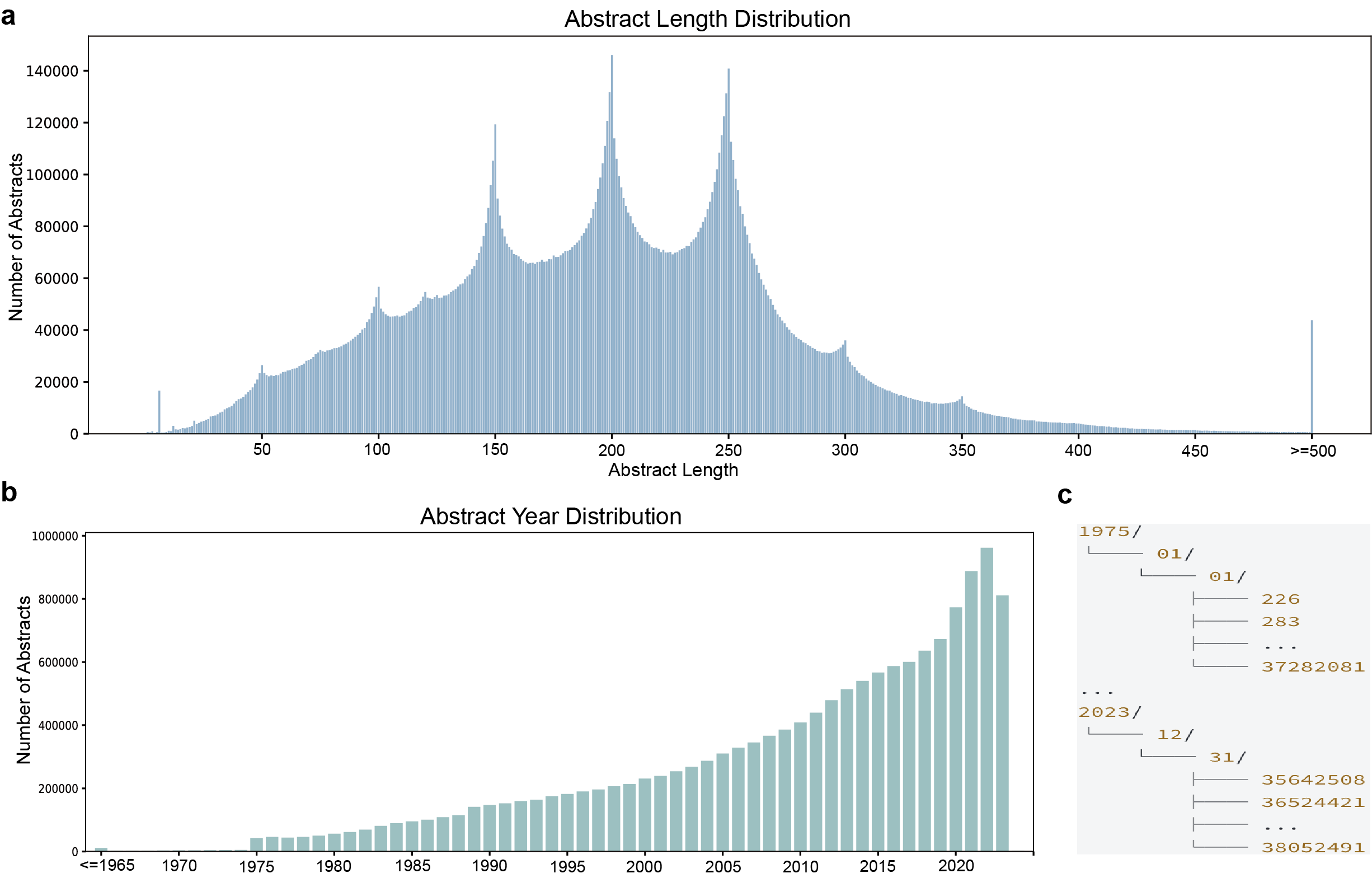}
    \caption{Data preprocessing and curation of PubMed abstracts. 
    Overview of the filtering and organization steps applied to over 20 million PubMed abstracts. 
    Abstracts were filtered by length and publication year, and then structured into a daily time series from 1975 to 2023. 
    \textbf{a}. Distribution of abstract lengths. The initial dataset displayed a long-tail distribution, with abstracts shorter than 100 words or longer than 500 words (aggregated for clarity) being infrequent and often containing uninformative content. 
    Only abstracts ranging from 100 to 300 words were retained. \textbf{b}. Publication year distribution. A marked increase in publications began in 1975. 
    Records from before 1965 were sparse and aggregated, while data from 2024 were excluded due to incompleteness. 
    The final dataset spans 1975-2023. 
    \textbf{c}. Daily time series structure. 
    A total of 10,014,314 abstracts were organized into a fine-grained daily time series from January 1, 1975, to December 31, 2023. Abstracts published on the same day were ordered by ascending PubMed ID to ensure intra-day consistency.}
\end{extendedfigure}

\begin{extendedfigure}
    \centering
    \includegraphics[width=1.0\textwidth]{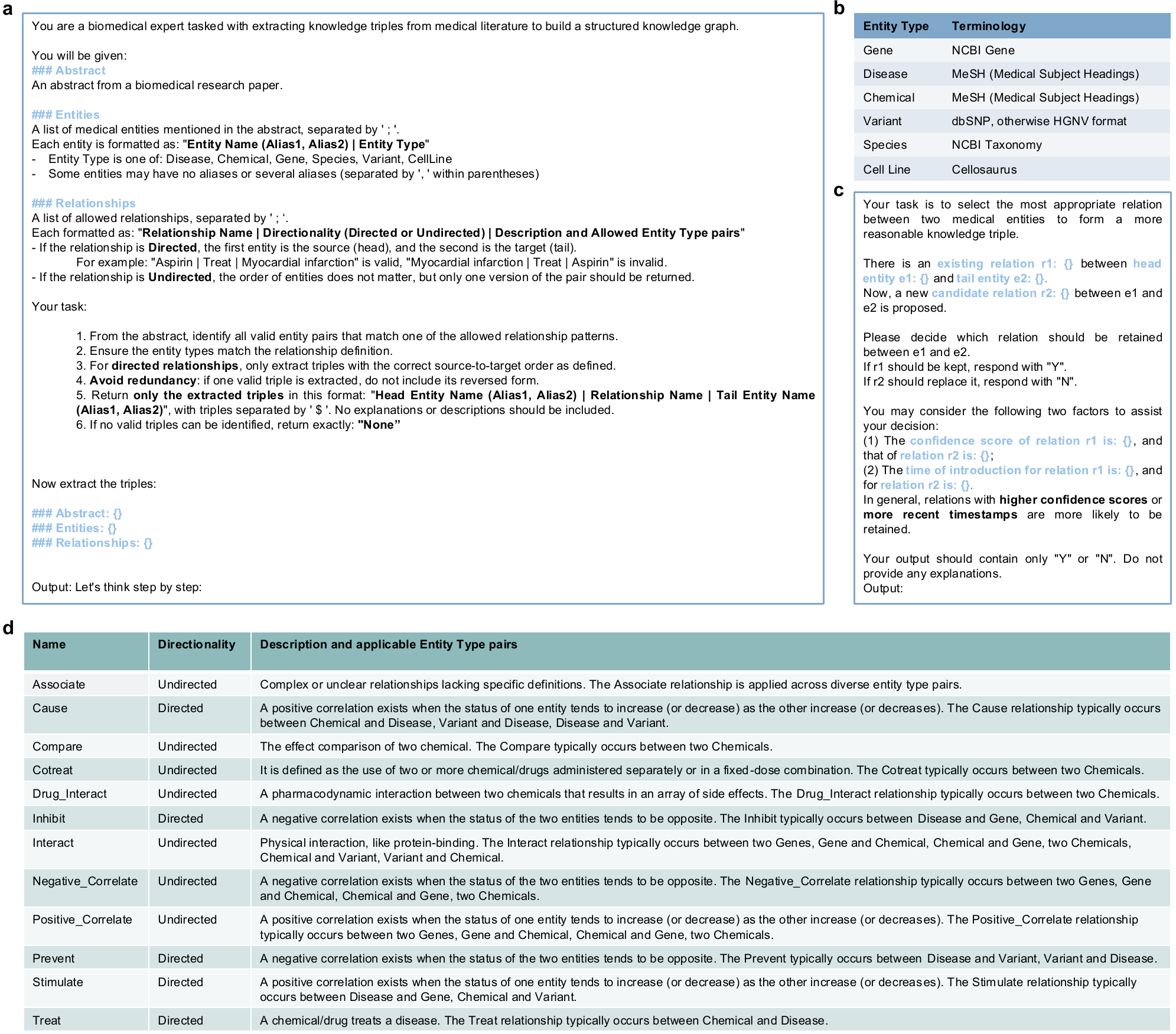}
    \caption{\textbf{a}. Prompt template for relation extraction. Given a biomedical abstract and its extracted entities, the Extractor Agent prompts the LLM to infer semantic relations between entity pairs using a predefined relation set and textual descriptions. 
    \textbf{b}. Reference terminologies for entity normalization. 
    Each biomedical entity type is mapped to a standard terminology: Gene (NCBI Gene), Disease and Chemical (MeSH), Variant (dbSNP or HGNV), Species (NCBI Taxonomy), and Cell Line (Cellosaurus). 
    \textbf{c}. Prompt design for relation conflict resolution. When conflicting relations exist between the same entity pair, the Constructor Agent prompts the LLM to select the most appropriate one based on confidence scores and timestamps. 
    \textbf{d}. Schema for predefined relation types. 
    The 12 core relation types---seven bidirectional and five unidirectional---are listed alongside their directionality, descriptions, and allowed entity-type combinations.}

\end{extendedfigure}

\begin{extendedfigure}
    \centering
    \includegraphics[width=1.0\textwidth]{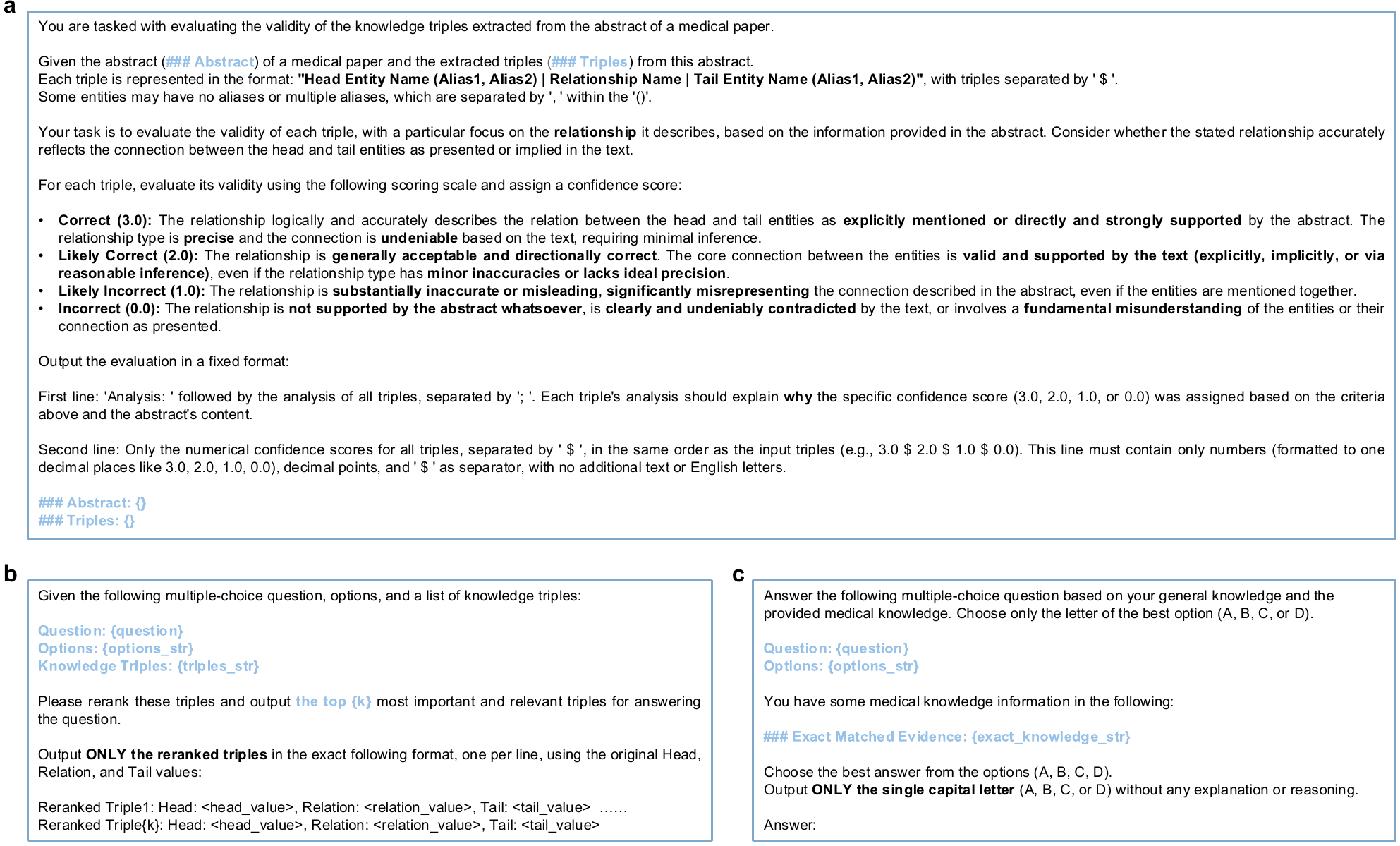}
    \caption{\textbf{a}. Prompt used for manual and LLM-based triple quality evaluation. 
    A standardized prompt presenting extracted relational triples was used to guide assessments by LLMs and domain-trained human annotators. 
    Each triple was scored on a four-level rubric ranging from 0.0 (Incorrect) to 3.0 (Correct). \textbf{b}. Prompt for subgraph reranking based on query relevance. 
    Given a query and its associated knowledge subgraph, the LLM is prompted to rank triples by their relevance to the question. 
    The top-k triples are retained to construct a refined subgraph used for downstream reasoning. \textbf{c}. Prompt for final answer generation. 
    In the reasoning phase, the question and the reranked subgraph are provided to the LLM to generate an answer. 
    This focused context reduces cognitive noise and improves response quality.}

\end{extendedfigure}

\end{document}